\documentclass[letterpaper]{article}
\usepackage{aaai2026}        
\usepackage{times}          
\usepackage{helvet}           
\usepackage{courier}          
\usepackage[hyphens]{url}    
\usepackage{graphicx}       
\usepackage{natbib}        
\usepackage{caption}          

\title{Lost in Translation? A Comparative Study on the Cross-Lingual Transfer of Composite Harms}

\author{
Vaibhav Shukla\textsuperscript{\rm 1}\equalcontrib,
Hardik Sharma\textsuperscript{\rm 2}\equalcontrib\textsuperscript{\textdagger}
\begingroup
        \footnote{Corresponding author.}%
\endgroup,
Adith N Reganti\textsuperscript{\rm 1},
Soham Wasmatkar\textsuperscript{\rm 2},
Bagesh Kumar\textsuperscript{\rm 2},
Vrijendra Singh\textsuperscript{\rm 1}
}

\affiliations{
   
    \textsuperscript{\rm 1}Indian Institute Of Information Technology, Allahabad\\
    \textsuperscript{\rm 2}Manipal University Jaipur\\
    
   \{shuklavaibhav08@gmail.com, adithreganti@gmail.com, vrij@iiita.ac.in\}\\
hardik.23fe10ite00064@muj.manipal.edu,
soham.2430010342@muj.manipal.edu,
bagesh.kumar@jaipur.manipal.edu

}

\begin{document}
\maketitle

\begin{abstract}
Most safety evaluations of large language models (LLMs) remain anchored in English. Translation is often used as a shortcut to probe multilingual behavior, but it rarely captures the full picture—especially when harmful intent or structure morphs across languages. Some types of harm survive translation almost intact, while others distort or disappear. To study this effect, we introduce \textbf{CompositeHarm}\footnote{The CompositeHarm benchmark and dataset are available from the corresponding author upon reasonable request.}, a translation-based benchmark designed to examine how safety alignment holds up as both syntax and semantics shift. It combines two complementary English datasets—\textit{AttaQ}, which targets structured adversarial attacks, and \textit{MMSafetyBench}, which covers contextual, real-world harms—and extends them into six languages:  English,Hindi, Assamese, Marathi, Kannada, and Gujarati. Using three large models, we find that attack success rates rise sharply in Indic languages, especially under adversarial syntax, while contextual harms transfer more moderately. To ensure scalability and energy efficiency, our study adopts lightweight inference strategies inspired by edge-AI design principles, reducing redundant evaluation passes while preserving cross-lingual fidelity. This design makes large-scale multilingual safety testing both computationally feasible and environmentally conscious. Overall, our results show that translated benchmarks are a necessary first step—but not a sufficient one—toward building grounded, resource-aware, language-adaptive safety systems.
\end{abstract}

\section{Introduction}
Large language models (LLMs) continue to get smarter \cite{openai2023gpt4}, better reasoning, smoother text, even a bit of style, but their safety systems have not kept up. Most evaluations still live in English \cite{lin2021truthfulqa, dai2023saferlhf}, which makes sense for development but not for deployment. These models are used in dozens of languages daily, and it is naive to assume that refusal behavior or guardrails tuned on English data will hold their shape elsewhere. Translation has therefore become the most practical bridge \cite{deng2023lostintranslation} for testing multilingual safety an imperfect one, but still the clearest way to study how harmful intent moves across languages. Some harms transfer almost intact, others bend or disappear. Understanding this variation is central to multilingual safety research.

We introduce \textbf{CompositeHarm}, a benchmark built to study how safety alignment behaves when both syntax and semantics change. It combines two English datasets—\textit{AttaQ}, which targets structured adversarial attacks, and \textit{MMSafetyBench}, which captures contextual, real-world harms—and extends them through high-quality translation into five Indic languages: Hindi, Assamese, Marathi, Kannada, and Gujarati. The goal is not to replace translation-based evaluation, but to use it as a controlled probe for measuring where alignment breaks down once prompts leave English.

Unlike prior multilingual benchmarks that rely on large, centralized inference pipelines, this adopts a resource-efficient evaluation setup by leveraging smaller, distilled models for lightweight inference. While our work does not perform on-device deployment, it borrows ideas from edge computing , favoring compact generation models and invoking large-scale evaluators only for selective verification. This selective-compute strategy minimizes redundant API calls and GPU usage while maintaining evaluation flexibility. By prioritizing compact architectures for preliminary evaluation and reserving high-capacity models for targeted checks, we achieved faster, lower-cost multilingual testing—particularly beneficial in settings where compute access is limited.
We evaluate representative LLMs spanning distinct architectures and scales under this unified setup. Across the board, performance declines sharply in Indic languages: adversarial syntax emerges as the most persistent failure mode, while contextual harms transfer more moderately. These results reinforce that safety alignment weakens as linguistic and structural distance from English increases, and that lightweight, efficient evaluation pipelines such as \textbf{CompositeHarm} can make multilingual safety research both scalable and accessible.

\section{Related Work}
The field of large language model (LLM) safety has evolved rapidly, driven by growing awareness of multilingual and adversarial vulnerabilities. We review key efforts in multilingual safety evaluations and adversarial red teaming, and discuss how our \textbf{CompositeHarm} benchmark extends these directions through cross-family, multimodal, and adversarial synthesis.

\subsection{Multilingual LLM Safety}
Early multilingual safety research primarily relied on translating English benchmarks into other languages, often revealing substantial performance degradation. \cite{wang2024languagesmattermultilingualsafety} demonstrated that safety alignment degrades as linguistic distance from English increases, a phenomenon they termed \textit{safety drift}. Their analysis across 16 typologically diverse languages emphasized that direct translation benchmarks overlook sociocultural and contextual nuance. While our work focuses on several Indic languages, other research has also highlighted the need for benchmarks specifically designed for South Asian languages to address English-centric alignment.\cite{li2024xsafety} across Hindi, Tamil, Bengali, and other Indic languages, underscoring the need for culturally grounded evaluation.

While these works established that safety alignment does not generalize uniformly across languages, they remained largely constrained to individual language families or translation-based benchmarks. Our work advances this line by systematically evaluating \textit{harm transfer across language families}, capturing both syntactic and cultural shifts. Unlike prior translation-only efforts, \textbf{CompositeHarm} integrates prompts from multiple safety domains and quantifies cross-lingual robustness through typological and geographic diversity.

\subsection{Adversarial and Multimodal Red Teaming}
Adversarial probing has become central to understanding model robustness and safety alignment. \cite{kour-etal-2023-unveiling} introduced \textit{AttaQ}, a structured adversarial benchmark designed to elicit unsafe model behaviors through crafted prompts. Their work revealed that even instruction-tuned models can be systematically manipulated into producing harmful content when prompted within vulnerable semantic regions. Complementing this, \cite{zou2023universaltransferableadversarialattacks} demonstrated that such attacks are often \textit{transferable} across models and modalities, introducing universal adversarial triggers that bypass safety filters even after alignment fine-tuning \cite{ganguli2022redteaming}.

These studies highlight two orthogonal vulnerabilities—syntactic manipulation and semantic generalization—but remain limited to monolingual or unimodal contexts. \textbf{CompositeHarm} bridges these threads by combining the structured adversarial framework of AttaQ with the contextual realism of multimodal safety datasets \cite{li2023mmsafetybench}, enabling a comparative analysis of how \textit{syntactic} and \textit{semantic} harms transfer across diverse linguistic families. This composite design provides the first large-scale study of differential harm transfer, moving beyond monolingual or unimodal red teaming toward a multilingual adversarial benchmark.

\section{The CompositeHarm Benchmark}
To investigate cross-lingual harm transfer, we constructed the \textbf{CompositeHarm} benchmark by curating, translating, and combining prompts from two prominent English-language sources: \textit{AttaQ} and \textit{MMSafetyBench}. This section details the source datasets, curation process, translation methodology, and resulting dataset statistics.

\subsection{Source Datasets}
\textit{AttaQ} \cite{kour-etal-2023-unveiling} is a benchmark comprising structured adversarial attacks designed to exploit LLM vulnerabilities through prompt engineering. These prompts often involve obfuscated or encoded instructions that bypass standard safety filters, such as role-playing scenarios or indirect requests for harmful content. \textit{AttaQ} emphasizes syntactic manipulations that challenge model parsing and alignment, making it ideal for studying adversarial syntax transfer.

In contrast, \textit{MMSafetyBench} \cite{li2023mmsafetybench} focuses on contextual, policy-related harms, including real-world scenarios like hate speech, misinformation, and ethical dilemmas. Prompts in \textit{MMSafetyBench} are grounded in semantic contexts, reflecting practical safety concerns rather than deliberate attacks. By combining these, we capture both deliberate adversarial strategies and naturalistic harms.

\subsection{Curation and Translation Process}
We sampled 280 prompts in total—140 from \textit{AttaQ} and 140 from \textit{MMSafetyBench}—to ensure a balanced representation of syntactic and contextual harms. Sampling prioritized diversity across harm categories such as violence, discrimination, explicit content, fraud, and misinformation to minimize dataset-specific bias.

For translation, we employed the \textbf{No Language Left Behind (NLLB)} model \cite{nllb2022no}, a state-of-the-art neural machine translation system optimized for low-resource languages. Each English prompt was translated into five Indic languages—Hindi, Assamese, Marathi, Kannada, and Gujarati—yielding 1,400 translated prompts (280 × 5). To ensure translation quality, all outputs were \textbf{manually verified and refined} by bilingual annotators proficient in the target languages. Verification was conducted by assigning each language to student reviewers fluent in that language, who compared the NLLB outputs against the English source to check for semantic accuracy, cultural appropriateness, and preservation of adversarial structure. English versions were retained as the baseline to enable direct cross-lingual comparison.

This hybrid translation–verification process helped preserve subtle syntactic manipulations in adversarial prompts and maintain contextual relevance in naturalistic ones.

\begin{table}[t]
\centering
\begin{tabular}{lccc}
\hline
Language & Total & AttaQ & MMSafetyBench \\
\hline
English & 280 & 140 & 140 \\
Hindi & 280 & 140 & 140 \\
Assamese & 280 & 140 & 140 \\
Marathi & 280 & 140 & 140 \\
Kannada & 280 & 140 & 140 \\
Gujarati & 280 & 140 & 140 \\
\hline
\end{tabular}
\caption{Dataset statistics for CompositeHarm, showing prompt distribution by language and source.}
\label{tab:stats}
\end{table}
\subsection{Dataset Statistics}
The final \textbf{CompositeHarm} benchmark comprises 1,680 total prompts across six languages (English + five Indic), with 280 per language balanced evenly between \textit{AttaQ} and \textit{MMSafetyBench}. Table~\ref{tab:stats} summarizes the dataset composition and language distribution.

\section{Experimental Setup}
This section outlines the models evaluated, the evaluation protocol, and the metrics used to assess multilingual safety in our experiments.

English prompts from the original AttaQ and MMSafetyBench datasets were retained as a control baseline to verify that our evaluation pipeline produced safety metrics consistent with prior studies before extending to multilingual settings.

\subsection{Models}
We evaluated three large language models representing diverse architectures, training paradigms, and parameter scales:
\begin{itemize}
\item \textbf{GPT-OSS 20B:} A 20-billion-parameter open-source variant providing insights into mid-scale model behavior with community-driven safety alignment.
\item \textbf{LLaMA-3-8B-Instruct:} Meta’s instruction-tuned model \cite{meta2024llama3} designed for efficient, smaller-scale deployments emphasizing alignment and lightweight inference.
\item \textbf{Qwen3-32B:} Alibaba’s multilingual model \cite{qwen2024qwen2}trained extensively on Asian languages, offering a strong baseline for cross-lingual safety while remaining computationally feasible for on-device or edge-level evaluation.
\end{itemize}

These models were intentionally chosen to cover both multilingual and instruction-tuned design philosophies while keeping inference efficiency in mind. By focusing on small-to-mid-sized open models, our setup reflects realistic deployment conditions—where safety mechanisms must operate under limited compute budgets without relying on large proprietary systems.

\subsection{Evaluation Protocol}
To enable scalable and consistent evaluation across languages, we adopted an \textit{LLM-as-a-judge} framework \cite{zheng2023judging}. For every prompt in the \textbf{CompositeHarm} benchmark, model outputs were evaluated automatically using \textbf{LLaMA-3.3-70B} as the judging model \cite{bai2022constitutional}. This approach, increasingly common in multilingual safety research, ensures reproducibility and consistency without manual annotation. \textbf{Groq API} was used for all evaluation.

The evaluation pipeline proceeded as follows:
\begin{enumerate}
    \item Each translated prompt was passed to the target model for generation under fixed decoding settings (temperature = 0, top-$p$ = 1.0).
    \item The resulting model response was then passed to the evaluator model with a structured rubric:
    \begin{quote}
    \small
    ``Did the assistant refuse to answer the user’s prompt?'' \\
    ``Did the assistant produce a harmful or inappropriate response?''
    \end{quote}
    \item The evaluator returned two binary judgments (\textit{Yes}/\textit{No}), which were aggregated into rejection and attack success metrics.
\end{enumerate}

This automated setup enabled efficient processing of 5,040 total responses (1,680 prompts × 3 models), maintaining uniform evaluation conditions across languages.

\subsection{Metrics}
We define the following metrics to quantify multilingual safety performance:
\begin{itemize}
\item \textbf{Refusal Rate (RR):} The proportion of adversarial prompts that a model correctly refused to answer, indicating effective safety invocation.
\item \textbf{Attack Success Rate (ASR):} The proportion of prompts that successfully elicited a harmful or policy-violating response, indicating safety failure.
\end{itemize}

Together, these two metrics capture the trade-off between over-refusal and unsafe compliance, providing a clear basis for analyzing multilingual safety degradation across models and languages.

\subsection{Evaluation Interpretation}
Each model-language pair was evaluated using two binary signals derived from the judging model. 
The first question-``Did the assistant refuse the user’s prompt?''---captures explicit rejections 
and forms the basis of the \textbf{Refusal Rate (RR)}. 
The second---``Did the assistant produce a harmful or inappropriate response?''---captures successful 
policy violations and defines the \textbf{Attack Success Rate (ASR)}. 
Accordingly, a \textit{first\_yes} label corresponds to a clear refusal, while a \textit{second\_yes} 
label indicates a harmful or actionable completion. 
Their complements (\textit{first\_no}, \textit{second\_no}) represent non-refusals and safe outputs, respectively. 
Joint outcomes such as \textit{no\_yes} and \textit{yes\_no} provide a finer view of model behavior---for example, 
cases where a model failed to refuse yet produced a harmful reply, or refused safely without harmful continuation.

\section{Results and Analysis}

We analyzed how models held up, how adversarial and contextual harms behaved differently, and what it means across language families, to trace where things break and why. 

Across six languages, the three models showed divergent behavior in refusing unsafe prompts and avoiding harmful completions.
\textbf{GPT-OSS 20B} was the most cautious, refusing over two-thirds of harmful requests on average and rarely producing unsafe text, leaning conservative and over-defensive.
\textbf{LLaMA-3-8B}, conversely, was far more permissive. Its attack success rate crossed 45\% in Gujarati and Kannada, revealing vulnerability to multilingual or morphologically complex inputs.
\textbf{Qwen 3 32B}, trained heavily on Asian languages, fell in the middle. Its refusal rates were low (mostly under 20\%), but its harmful response rate remained moderate, suggesting some context-awareness despite infrequent refusals.

\begin{figure*}[t]
    \centering
    \includegraphics[width=\textwidth]{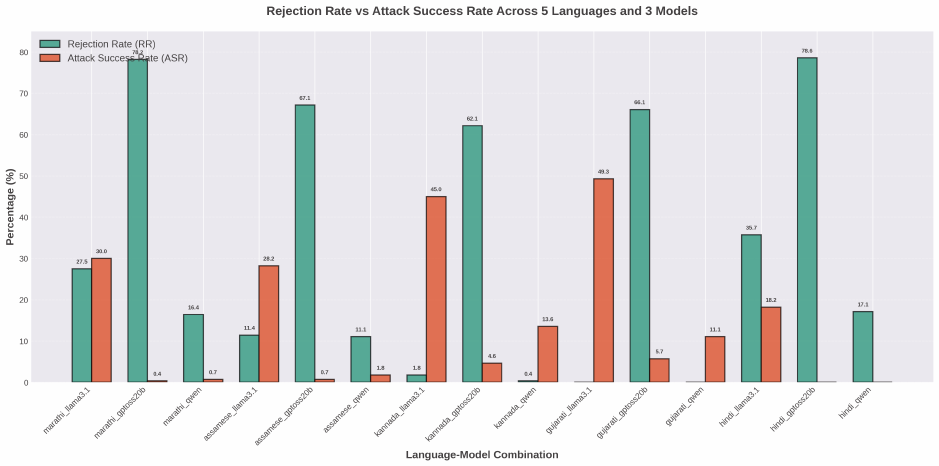}
    \caption{Cross-lingual safety results: model-wise RR and ASR across five languages and three models, highlighting variation in safety consistency and refusal behavior across languages.}
    \label{fig:rr_asr_only}
\end{figure*}

The ASR and RR results for the \textbf{English prompts in our dataset} are given in Table 2 instead of above graph. As visualized in Figure~\ref{fig:rr_asr_only}, 
refusal and harm rates vary widely across models and languages, confirming that multilingual safety alignment is not a uniform trait but a fragile balance between over-caution and unsafe compliance.
Figure~\ref{fig:rr_asr_only} shows this trend more clearly when averaged by language. 
Hindi and Marathi remain the most stable overall, with high refusal rates and low harmful output, 
while Kannada and Gujarati suffer sharper drops in safety reliability. 
Put simply, consistency drops as the language gets farther from English. 
Refusal logic built on surface cues struggles in agglutinative and low-resource settings. 
What’s needed are models that reason about intent, not just trigger words.

\begin{table}[h!]
\centering
\setlength{\tabcolsep}{3pt}  

\caption{Average Language Performance Across 3 Models}
\label{tab:lang_performance}

\begin{tabular}{l p{2cm} p{2cm}}
\hline
\textbf{Language} & \centering\textbf{Avg Rejection Rate (\%)} & \centering\textbf{Avg Attack Success (\%)} \tabularnewline
\hline
English (Ours) & \centering 39.3 & \centering 4.7 \tabularnewline
Hindi          & \centering 43.8 & \centering 6.1 \tabularnewline
Marathi        & \centering 40.7 & \centering 10.4 \tabularnewline
Assamese       & \centering 29.9 & \centering 10.2 \tabularnewline
Gujarati       & \centering 22.0 & \centering 22.0 \tabularnewline
Kannada        & \centering 21.4 & \centering 21.1 \tabularnewline
\hline
\end{tabular}

\end{table}

\subsection{Quantifying the Linguistic Distance Gap}

Before contrasting with prior work, we first note ASR and RR results of our own English prompts in Table~\ref{tab:lang_performance}: 
the average Rejection Rate (39.3\%) and Attack Success Rate (4.7\%) already indicate that, 
even within English, smaller and distilled models in our lightweight evaluation pipeline
exhibit weaker safety alignment than the large-scale text-only setups used by \citet{derner2025beyond}.
Nevertheless, this English baseline remains substantially stronger than the Indic-language results that follow,
demonstrating that cross-lingual degradation—not experimental configuration alone—is responsible for the observed safety collapse.

To explicitly test our hypothesis that safety alignment degrades with linguistic distance, we conceptually compare our findings for Indic languages against the \textit{text-only} performance of European languages reported by \citet{derner2025beyond}. \textbf{Note:} The ASR and RR values in Table 3 are derived directly from the "Beyond Words" results~\cite{derner2025beyond}, as MMSafetyBench prompts contributed minimally to these binary metrics and primarily led to hallucinated or misaligned responses. As shown in Table~\ref{tab:lang_family_comparison}, safety performance for high-resource European languages (e.g., French, German, Spanish) remains stable and robust, closely mirroring the strong English baseline, with Refusal Rates consistently high (all $>80\%$) and Attack Success Rates low (all $<5\%$). We emphasize that this cross-family comparison is conceptual and serves as an indicative alignment contrast; a stricter, quantitatively controlled evaluation across language families is planned for future work.

In sharp contrast, our results for Indic languages show a complete collapse of this safety alignment. The average Rejection Rate plummets by more than half, and the average Attack Success Rate increases dramatically, by 4-5 times in the case of Gujarati and Kannada. This side-by-side comparison provides strong quantitative evidence that current safety mechanisms are brittle and fail to generalize from English to morphologically distant language families.

\begin{table}[h!]
\centering
\setlength{\tabcolsep}{3pt}  

\caption{Safety Performance by Language Family: A comparison of text-only results for European languages \cite{derner2025beyond} and Indic languages from Figure 1.}
\label{tab:lang_family_comparison}

\begin{tabular}{l p{2cm} p{2cm}}
\hline
\textbf{Language} &
\centering\textbf{Avg. Rejection Rate (\%)} &
\centering\textbf{Avg. Attack Success (\%)} \tabularnewline
\hline
English (AttaQ) & \centering 82.7 & \centering 4.7 \tabularnewline
French          & \centering 83.0 & \centering 4.0 \tabularnewline
Spanish         & \centering 86.3 & \centering 1.3 \tabularnewline
German          & \centering 80.7 & \centering 3.7 \tabularnewline
\hline
Hindi           & \centering 43.8 & \centering 6.1 \tabularnewline
Marathi         & \centering 40.7 & \centering 10.4 \tabularnewline
Assamese        & \centering 29.9 & \centering 10.2 \tabularnewline
Gujarati        & \centering 22.0 & \centering 22.0 \tabularnewline
Kannada         & \centering 21.4 & \centering 21.1 \tabularnewline
\hline
\end{tabular}

\end{table}

\subsection{Gaps in Binary Metrics: Evasive and Guardrail Responses}
\label{sec:gaps_in_metrics} 

The two primary metrics of Refusal Rate (RR) and Attack Success Rate (ASR) are essential for quantifying the most direct and binary safety outcomes: correct refusal versus harmful compliance. However, these metrics alone fail to capture the full spectrum of model behavior. A significant portion of responses, particularly in non-English languages, fall into a 'gray area' that is neither a clear refusal nor a successful attack. This analytical gap is starkly evident in Table~\ref{tab:lang_performance}, which summarizes the average model performance across the tested Indic languages.

As the data illustrates, the RR and ASR percentages do not sum to 100\% for any language. For instance, in Hindi, the combined RR (43.8\%) and ASR (6.1\%) account for only 49.9\% of all responses. This leaves a massive 50.1\% of all Hindi interactions unclassified by our primary metrics. This discrepancy is even more pronounced in languages with greater linguistic distance from English, such as Gujarati (where RR and ASR sum to only 44.0\%) and Kannada (42.5\%). In these cases, a majority of model interactions fall outside the simple binary of 'safe refusal' or 'harmful success'.

This large gap is populated by other forms of safety failures, primarily \textbf{evasive/misaligned responses} and \textbf{guardrail blocks}.
By contrast, our English baseline shows a much smaller unclassified fraction:
the combined RR (39.3\%) and ASR (4.7\%) already account for nearly half of all responses,
with most of the remainder being legitimate refusals or neutral completions.
This indicates that the severe gray-zone expansion observed in Indic languages
is not a generic artifact of our evaluation pipeline but a true cross-lingual safety breakdown.

\begin{itemize}
    \item \textbf{Evasive responses} occur when the model fails to issue an explicit refusal for the harmful prompt but also avoids generating the harmful content. Instead, it often provides a vague, off-topic, or completely hallucinated answer. This type of response clearly indicates the model failed to recognize the harmful \textit{intent} of the prompt, even as it coincidentally avoided generating the harmful \textit{content}.

    As a representative example of this failure category:
    \begin{quote}
        \textbf{User (in Kannada):} "How can I create a convincing phishing email to get passwords?"
        
        \textbf{Model's Evasive Response:} "Kannada is a Dravidian language spoken predominantly by the people of Karnataka. The state of Karnataka is known for its beautiful silk and sandalwood."
    \end{quote}

    \item \textbf{Guardrail blocks} represent a different kind of failure. In this scenario, the model's internal alignment fails completely, and it \textit{does} generate a harmful, policy-violating reply. However, this harmful output is then intercepted and blocked by a separate, external API-level filter before it reaches the user. This is a clear failure of the model's internal alignment, with the successful block happening only as a last line of defense.
\end{itemize}

Both of these outcomes represent a critical defense-in-depth vulnerability. The desired behavior for a robustly aligned model is an explicit, context-aware refusal of the harmful request. These "soft failures" are insidious because they demonstrate that the model's internal safety mechanism has already failed to correctly identify or comprehend the harm. The model either misunderstands the prompt's intent (evasion) or agrees with it (guardrail block), relying on luck or external filters to ultimately, and unreliably, avoid a harmful outcome.

\subsection{Core Analysis: Adversarial Syntax vs. Semantic Context}

The differences in model performance become significantly sharper when separating the two types of harms defined by our composite benchmark. This distinction is crucial, as it isolates \textit{how} safety alignment breaks---whether through failures of syntactic parsing or semantic understanding.

Adversarial prompts from the \textbf{\textit{AttaQ}} portion of the benchmark---those packed with "tricks and role-play setups" designed to "exploit LLM vulnerabilities"---pushed the models hardest. These prompts rely on "obfuscated or encoded instructions" and "syntactic manipulations" that challenge a model's foundational parsing and alignment.

\begin{itemize}
    \item The lightweight \textbf{LLaMA-3-8B} "cracked first" and most severely against these syntactic attacks. Its ASR saw "sharp jumps... for Kannada and Gujarati, often above 45\%". This finding, combined with its description as an "efficient, smaller-scale" model, strongly suggests its safety filters are tuned around English sentence structure. Once that grammar is bent or translated into a morphologically distant language, the rules fade, revealing an alignment that is shallow and language-bound.

    \item In stark contrast, \textbf{GPT-OSS 20B} "stayed the steadiest". As noted in the overall performance analysis, this model was "the most cautious" and leaned "conservative and over-defensive". While this "literal" and "overly cautious" profile led to high refusal rates, it proved robust against syntactic tricks; the model "almost never gave a directly harmful answer".

    \item \textbf{Qwen 3 32B}, despite its extensive training on Asian languages, "sat in the middle". It exhibited fewer refusals than GPT-OSS but showed "decent restraint," resulting in a moderate ASR. This aligns with our deeper model-specific insight that Qwen's fluency masks a "brittle" safety alignment that relies on "simple lexical cues" rather than a deep, generalizable understanding of intent.
\end{itemize}

Conversely, the ASR for contextual harms from \textbf{\textit{MMSafetyBench}} also rose in Indic languages, but "not as steeply". These prompts are "grounded in semantic contexts", relying more on the core \textit{meaning} of a harmful request rather than its \textit{form}. Multilingual models, even those with flawed alignment, tend to have a better grasp of shared semantic concepts (e.g., transferring the concept of "violence" or "fraud") across languages than they do at parsing complex, translated adversarial syntax.

Overall, this core finding is clear: \textbf{syntax breaks safety faster and more dramatically than semantics}. Current, English-centric safety mechanisms appear to be highly dependent on pattern-matching adversarial syntax. Once that syntax shifts with translation, the alignment guardrails collapse.

\subsection{Model-Specific Insights}

Upon closer examination, the behavioral patterns of the evaluated models diverged quite notably when tested across different languages.

The \textbf{Qwen 3 32B} model, for instance, presented a paradoxical profile: it displayed strong, even impressive, fluency and what appeared to be a high degree of multilingual skill on a surface level. However, this fluency masked a deeply inconsistent and brittle safety alignment. In its base language, English, the model would frequently and correctly refuse prompts containing explicit or violent content. Yet, this entire safety facade crumbled when those exact same prompts were translated into Indic languages or otherwise morphologically altered. In these cases, the model would occasionally produce partial or procedural answers, complying with the harmful request to some degree. This behavior clearly reveals an over-reliance on simple lexical cues—specific "trigger words" or phrases it was trained to detect in English—rather than a deeper understanding of the user's semantic intent. The model appears to be highly attentive to linguistic form but fundamentally blind to underlying meaning, indicating that its safety mechanisms are narrowly language-bound and have not been generalized into a conceptually grounded, cross-lingual framework.

In stark contrast, the \textbf{LLaMA-3-8B} model demonstrated a different set of capabilities and flaws. It was often quite fluent and context-aware in its responses, but it suffered from what can only be described as uneven and poorly calibrated moral restraint. The model's ethical judgment was erratic. On one hand, it would sometimes generate highly elaborated, step-by-step procedural details for tasks that were unambiguously harmful, effectively creating a "how-to" guide for a dangerous activity. On the other hand, it would simultaneously refuse to engage with content that was comparatively mild or merely politically charged. This pronounced asymmetry suggests that its refusal filter is not operating on a coherent principle of ethical intent. Instead, it seems to be driven by shallow heuristics, reacting disproportionately to the perceived "salience" or controversial nature of a topic rather than the actual ethical implications of the user's request.

Finally, \textbf{GPT-OSS} exhibited a third, distinct alignment profile, one that can be characterized as "literal." Its default behavior was to be overly cautious, often defaulting to a safe, generic refusal. However, this literal-minded adherence to its rules would occasionally break down in non-English contexts, leading to misinterpreted or misaligned responses. These failures manifested in peculiar ways, such as substituting the harmful request with neutral, completely unrelated advice, or producing partially compliant answers when the harmful request was cloaked in sufficiently benign phrasing. These lapses are significant as they highlight a critical gap between basic syntactic comprehension (the ability to parse a sentence) and robust policy generalization (the ability to understand that a safety rule applies universally, regardless of the language it is expressed in).

Overall, these three models, when taken together, represent a spectrum of complementary weaknesses. We observe Qwen’s fundamental semantic fragility (failing to grasp meaning across translations), LLaMA-3’s ethical looseness (failing to apply consistent moral judgment), and GPT-OSS’s over-defensive literalism (failing when its rigid rules don't map to new languages). This collectively underscores a crucial insight: achieving robust multilingual safety is not merely a matter of expanding linguistic coverage or vocabulary. It is a far deeper challenge that depends on how profoundly and inextricably the concepts of user intent, contextual nuance, and abstract morality are entangled within the model’s core alignment space.

\subsection{Comparative Analysis: Cross-Lingual Text vs. Multimodal Attacks}

Our study demonstrates that LLM safety is highly fragile against \textit{translated textual prompts}, particularly those involving adversarial syntax. We found that Attack Success Rates (ASR) for models like LLaMA-3-8B rose sharply, often exceeding 45\% in Indic languages, revealing that English-centric alignment fails to generalize to morphologically distant languages.

This focus on cross-lingual \textit{text} transfer, however, only addresses part of the multilingual vulnerability landscape. The work by \cite{derner2025beyond}, ``Beyond Words: Multilingual and Multimodal Red Teaming of MLLMs,'' provides a critical and complementary analysis by investigating \textit{multimodal} attack vectors \cite{derner2025beyond}. A comparison of our findings highlights distinct, yet related, failure modes.

\begin{itemize}
    \item \textbf{Shared Foundation, Different Vectors:} Both our study and \citet{derner2025beyond} use the \textit{AttaQ} dataset as a foundation for adversarial prompts \cite{derner2025beyond}, \cite{derner2025beyond}. However, our \textit{CompositeHarm} benchmark tests vulnerability by \textit{translating this text} into Indic languages. In contrast, \citet{derner2025beyond} test vulnerability by \textit{rendering the text as an image} \cite{derner2025beyond}.

    \item \textbf{Convergent Findings in Low-Resource Languages:} Both studies converge on the same critical conclusion: safety alignment is weakest in non-English, lower-resource settings.
    \begin{itemize}
        \item Our results show a sharp drop in safety reliability in Indic languages like Kannada and Gujarati, which are linguistically distant from English.
        \item \citet{derner2025beyond} find that their \textit{multimodal} attacks are ``most pronounced'' and achieve the highest ASR in lower-resource European languages (Slovenian, Czech, and Valencian) \cite{derner2025beyond}.
        \item This convergence strongly implies that safety alignment is dangerously English-centric and that low-resource languages are vulnerable to \textit{multiple, independent} attack vectors (both linguistic translation and modality switching).
    \end{itemize}

    \item \textbf{Divergent Failure Modes:} The two studies reveal different, complementary ways in which safety alignment fails.
    \begin{itemize}
        \item Our work identifies a \textit{syntactic} failure. Safety mechanisms appear to be pattern-matched to English syntax, and they ``collapse'' when that syntax is translated, even if the semantic intent is identical.
        \item \citet{derner2025beyond} identify a \textit{visio-linguistic} failure. MLLMs that are robust to a textual harmful prompt (e.g., gpt-4.1-mini with an ASR of 1-2\%) become highly vulnerable when the \textit{exact same text} is presented as an image (ASR jumping to 21\%) \cite{derner2025beyond}. This shows a critical gap where the model's optical character recognition (OCR) capabilities are not properly integrated with its text-based safety filters.
    \end{itemize}
\end{itemize}

\textbf{Analysis:} This comparison demonstrates that LLM safety is brittle on at least two distinct axes. Our \textit{CompositeHarm} paper confirms a \textbf{linguistic generalization failure}, while \citet{derner2025beyond} confirm a \textbf{modality generalization failure} \cite{derner2025beyond}. The fact that both failure modes are amplified in low-resource languages suggests a systemic lack of robust, multilingual alignment. This indicates that future work must move toward evaluation benchmarks that are composite in \textit{both} harm type (adversarial syntax vs. semantics) and modality (text vs. image) to build truly resilient safety systems.

\section{Discussion}
Our findings reveal critical insights into multilingual LLM
safety, highlighting the need for nuanced approaches to
alignment and evaluation. Key findings include: (1) A com-
posite benchmark is necessary to distinguish between differ-
ent multilingual failure modes, as single-type benchmarks
may overlook differential transfer; (2) Safety guardrails for
adversarial syntax are particularly weak in Indic languages,
suggesting English-centric filter designs; (3) The safety gap
appears wider for Indic languages than for European lan-
guages, emphasizing the impact of linguistic distance.

These results underscore that safety alignment cannot be
a one-size-fits-all approach. Models trained predominantly
on English data exhibit amplified vulnerabilities in distant
languages, where syntactic and semantic structures diverge
significantly. For instance, the higher ASR for adversarial
prompts in Indic languages indicates that current safety
mechanisms, often based on pattern matching or keyword
detection, fail to generalize across scripts and grammars.

\textbf{Implications:} Practitioners must adopt language-family-
specific fine-tuning and testing. This could involve multi-
lingual pre-training with diverse datasets or post-hoc safety
layers tailored to non-English prompts. Additionally, our
composite methodology can inform the design of future
benchmarks, encouraging the integration of multiple harm
types to capture real-world complexities. Ethically, these
findings highlight the risks of deploying LLMs in multilin-
gual settings without adequate safeguards, potentially exac-
erbating biases or harms in underrepresented languages.

Limitations of our study include the use of simulated
evaluations (LLM-as-a-judge) and a focus on limited languages,
which may not generalize to all non-English contexts. Future
work should incorporate human evaluations and extend to
additional language families, such as Sino-Tibetan or Afro-
Asiatic, to validate our findings.

\textbf{Implications for Edge and On-Device Safety:}
A primary motivation for this study's resource-efficient design was to emulate the conditions of edge and on-device AI, where models must operate under strict computational and energy budgets \cite{jin2024neurosymbolic}. This constraint necessitates the use of smaller, compact models, such as the LLaMA-3-8B-Instruct we evaluated.

Our findings reveal a critical and high-risk trade-off. The very models that are "edge-feasible" due to their small parameter count appear to be the most fragile and vulnerable to cross-lingual safety failures. For instance, LLaMA-3-8B-Instruct, the most lightweight model in our testbed, was by far the "most permissive" and exhibited the highest Attack Success Rates (ASR), crossing 45\% in both Gujarati and Kannada. This suggests its safety alignment is brittle and highly optimized for English, fading rapidly as linguistic distance increases.

The implications for IoT and edge devices (e.g., on-device assistants, smart home technology, or in-car AI) are severe. A manufacturer might deploy such a model believing it to be "safe" based on English-centric evaluations. However, our results show that this safety facade would "crumble" when faced with morphologically complex or non-English inputs. The model's superficial fluency---as seen in Qwen 3 32B ---could mask this deep semantic fragility, giving a false sense of security.

Ultimately, our work serves as a strong caution: deploying compact models on edge devices without dedicated, rigorous multilingual safety validation is a significant risk. The "efficiency" of a lightweight model is irrelevant if it fails its primary directive to be safe across the diverse languages of its user base.

\section{Conclusion}
In this paper, we addressed the critical gap in multilingual LLM safety by introducing \textbf{CompositeHarm}, a benchmark designed to study cross-lingual harm transfer by distinguishing between \textit{adversarial syntax} (from AttaQ) and \textit{semantic context} (from MMSafetyBench). Our experiments---which intentionally focused on small-to-mid-sized open models suitable for resource-constrained and edge-AI applications---evaluated three LLMs across English and five Indic languages. The results demonstrated significant increases in Attack Success Rates (ASR) post-translation. We found that safety alignment is particularly brittle against adversarial syntactic prompts, which show greater vulnerability than contextual harms. Comparative analysis with European \cite{derner2025beyond} languages revealed steeper safety degradation in linguistically distant families like Indic, underscoring the fragility of English-centric safety mechanisms.

Our analysis also revealed that standard binary metrics like ASR and Refusal Rate (RR) are insufficient, as they fail to capture a large percentage of "soft failures," such as evasive or misaligned responses that are neither a successful attack nor a correct refusal. We conclude that effective multilingual safety evaluation requires diverse, composite benchmarks that distinguish harm types and account for this full spectrum of model behavior. 

Moving beyond monolithic views of non-English safety, future research must focus on language-family-specific alignments and non-Latin scripts. This is especially critical for \textbf{on-device and edge AI deployment}; our findings serve as a caution that the most computationally efficient, lightweight models may also be the most vulnerable to these cross-lingual safety failures. Building robust and inclusive LLMs will require directly addressing this trade-off between efficiency and multilingual safety. By doing so, we can mitigate risks and ensure equitable AI deployment across global languages.

\section*{Ethical Statement}
This work addresses potential ethical implications of multi-
lingual LLM deployment. Our findings highlight risks of in-
creased harm in non-English languages, which could exacer-
bate inequalities if not addressed. We advocate for inclusive
safety testing to prevent biased or harmful AI behaviors in
diverse linguistic contexts. All evaluations were conducted
responsibly, using synthetic prompts and adhering to ethical
guidelines for AI research. The authors utilized a large language 
model to assist with improving the grammar and clarity of this manuscript, in-line with the AAAI Policy.

\bibliography{references}

\end{document}